\documentclass[letterpaper]{article} % DO NOT CHANGE THIS
\usepackage{aaai25}  % DO NOT CHANGE THIS
\usepackage{times}  % DO NOT CHANGE THIS
\usepackage{helvet}  % DO NOT CHANGE THIS
\usepackage{courier}  % DO NOT CHANGE THIS
\usepackage[hyphens]{url}  % DO NOT CHANGE THIS
\usepackage{graphicx} % DO NOT CHANGE THIS
\usepackage{multirow}
\usepackage{makecell}
\usepackage{natbib}  % DO NOT CHANGE THIS AND DO NOT ADD ANY OPTIONS TO IT
\usepackage{caption} % DO NOT CHANGE THIS AND DO NOT ADD ANY OPTIONS TO IT
\usepackage{algorithm}
\usepackage{algorithmic}

\usepackage{booktabs}
\usepackage{subcaption}
\usepackage{adjustbox}
\usepackage[disable,textsize=tiny]{todonotes}
\newcommand{\Eugene}[1]{\todo[backgroundcolor=red]{#1}}

\newcommand{\TheTMan}[1]{\todo[backgroundcolor=teal]{#1}}

\newcommand{\sam}[1]{\todo[backgroundcolor=green]{#1}}

\newcommand{\Z}[1]{\todo[backgroundcolor=yellow]{#1}}
\usepackage{newfloat}
\usepackage{listings}
\DeclareCaptionStyle{ruled}{labelfont=normalfont,labelsep=colon,strut=off} % DO NOT CHANGE THIS
\floatstyle{ruled}
\newfloat{listing}{tb}{lst}{}
\floatname{listing}{Listing}
\title{Video Game Level Design as a Multi-Agent Reinforcement Learning Problem}
\author{
Sam Earle\textsuperscript{\rm 1}, Zehua Jiang\textsuperscript{\rm 1}, Eugene Vinitsky\textsuperscript{\rm 1}, Julian Togelius\textsuperscript{\rm 1, \rm 2}
}
\affiliations{
    \textsuperscript{\rm 1}New York University\\
    \textsuperscript{\rm 2}University of Skövde\\
    sam.earle@nyu.edu
}

\usepackage{bibentry}
\begin{document}

\maketitle

\begin{abstract}
% An advantage of PCGRL is that it does not require example human-designed levels to produce high-quality levels. 
% Existing PCGRL research focuses on training single generator agents, and exploring the controllability and solvability of generated artifacts. 
Procedural Content Generation via Reinforcement Learning (PCGRL) offers a method for training controllable level designer agents without the need for human datasets, using metrics that serve as proxies for level quality as rewards. 
Existing PCGRL research focuses on single generator agents, but are bottlenecked by the need to frequently recalculate heuristics of level quality and the agent's need to navigate around potentially large maps.
% Existing PCGRL research focuses on single generator agents, but these systems often suffer from the need for frequent metric recalculations \sam{I don't think the reader of this abstract will know what a frequent metric recalculation is. I think they need some context here} and lack the ability to effectively adapt to or build upon existing structures \sam{Also not clear what this means}.
% We extend PCGRL by implementing \sam{I find this phrasing somewhat confusing. What does it mean to implement a multi-agent system. Maybe we should just cut this sentence and place the "more efficient and adaptive generation after the next sentence. It's also the case that I don't know what "efficient and adaptive" generation means, so maybe we just cut this entirely.}and evaluating a multi-agent system, where multiple generative agents collaboratively construct levels, leading to more efficient and adaptive generation.
By framing level generation as a multi-agent problem, we mitigate the efficiency bottleneck of single-agent PCGRL by reducing the number of reward calculations relative to the number of agent actions.
We also find that multi-agent level generators are better able to generalize to out-of-distribution map shapes, which we argue is due to the generators' learning more local, modular design policies.
We conclude that treating content generation as a distributed, multi-agent task is beneficial for generating functional artifacts at scale.
% We further explore specialization among diverse agents, observing how agents can collaborate, adopting locally-suboptimal, but ultimately complementary design strategies to result in globally optimal levels.
% different reward schemes?
% freezies --> maximum adaptivity?

% SECOND ATTEMPT AT ABSTRACT BELOW, WRITTEN BY JULIAN WITHOUT LOOKING AT THE ABOVE ABSTRACT. USE OR MIX AS YOU SEE FIT
% eyyy turn that shit down

% Reinforcement learning can be used to generate game levels, and similar content for training and simulation, through training level-designing agents. This has the advantage of resulting in compute-efficient level generation, but the disadvantage of the training being computationally expensive. The main reason for the slow training is that the level-designing agents require dense rewards, and reward calculation is expensive, which is in turn because estimating the quality of a level is nontrivial. We hypothesize that we can make training more efficient by turning procedural content generation into a multi-agent problem, so that multiple agents can share a single reward function calculation. This is because several agents can take actions simultaneously, and the reward calculation reflects all of their changes. We perform experiments with several different types of 2D game levels, and observe significant performance gains given the same number of reward function calculations. We analyze these results in some details and discuss the potential further advantages of a multi-agent formulation of PCGRL.

\end{abstract}

% Uncomment the following to link to your code, datasets, an extended version or similar.
%
\begin{links}
\link{Code}{https://github.com/smearle/pcgrl-jax}
% \link{Datasets}{https://aaai.org/example/datasets}
% \link{Extended version}{https://aaai.org/example/extended-version}
\end{links}

% \begin{teaserfigure}
% \centering
% \includegraphics[draft,width=.3\linewidth]{figs/overview.png}
% \caption{Multi-Agent PCGRL }
% \label{fig:overview}
% \end{teaserfigure}

%%% The next command prints the information defined in the preamble.

%%%%%%%%%%%%%%%%%%%%%%%%%%%%%%%%%%%%%%%%%%%%%%%%%%%%%%%%%%%%%%%%%%%%%%%%

\section{Introduction}

%SOME NEW JULIAN TEXT BELOW

Level design is a crucial part of video game development, and much work has gone into finding algorithmic means of generating or assisting in the design of video game levels. Generally, level generation is considered a form of Procedural Content Generation (PCG). This includes both level generation during gameplay, as is common in roguelikes and related game genres, and design assistance during game development. In addition to games for entertainment purposes, level generation is important for serious games (e.g. training simulations), as well as game-based AI benchmarks and training environments.

Many different methods have been developed for different forms of PCG, and in particular for level generation. % The method menagerie includes evolutionary algorithms, grammar expansion, constraint satisfaction, fractals, pattern substitution, supervised learning, and, more recently, reinforcement learning. 
In Procedural Content Generation via Reinforcement Learning (PCGRL, \citealt{pcgrl}), level-designing agents are trained to generate levels, and rewarded for producing designs that satisfy some reward function. This method offers fast generation times and a high degree of controllability, but requires long and costly training runs. This is because of the need for frequent, costly reward calculations, typically involving the computation of global shortest paths. Unlike most RL environment dynamics which are strictly local, these path-length computations have complexity $O(N^2)$, where $N$ is the maximum width of a given map.

\begin{figure*}
\includegraphics[trim={0 450 0 0},clip,width=\linewidth]{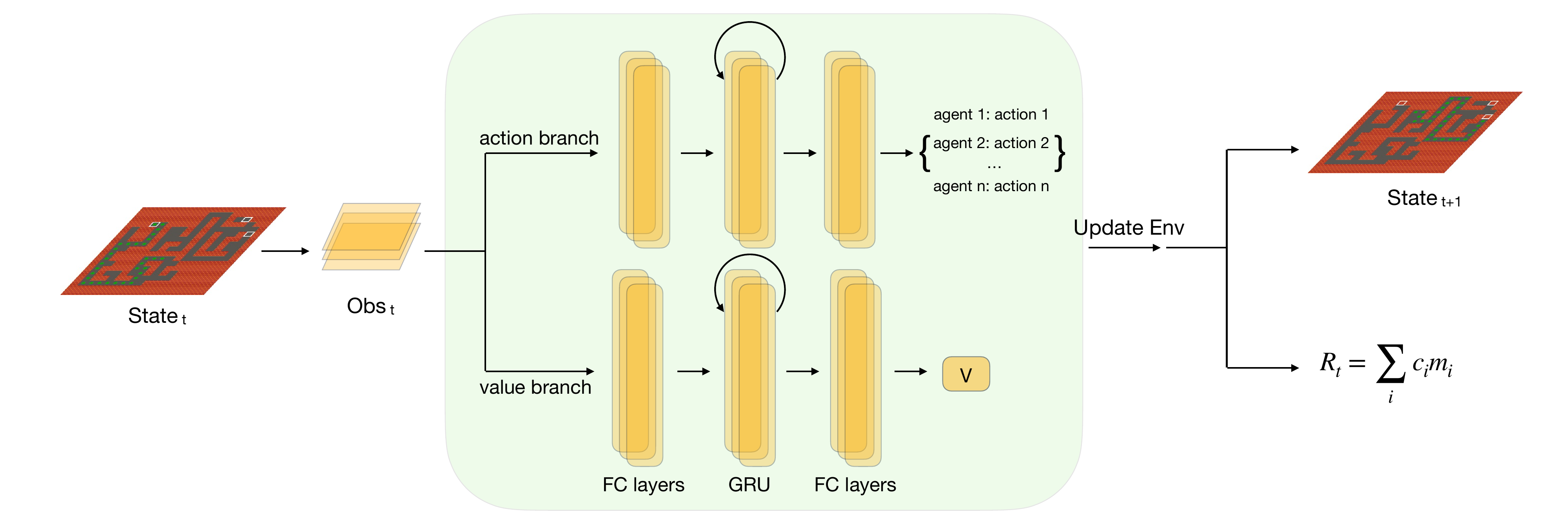}
\caption{In multi-agent PCGRL, agent actions are taken in parallel, reducing the number of reward computations required relative to level edits, while maintaining a per-agent dense reward scheme, and fostering collaboration between agents. Reward $R_t$ is computed according the weighted sum of heuristic scores $m_i$. \sam{was it intended to apply that there's a single global value that comes out of this network? If so, just ignore this but do clarify that in the caption.}}
\label{fig:overview}
\end{figure*}

\begin{table*}[t]
\centering
\adjustbox{max width=\textwidth}{%
% \scriptsize
\begin{tabular}{lllllllll}
\toprule
 & \multicolumn{8}{c}{mean episode reward} \\
map shape & \multicolumn{4}{c}{fixed} & \multicolumn{4}{c}{random} \\
map width & 8 & 16 & 24 & 32 & 8 & 16 & 24 & 32 \\
\toprule
n. agents &  &  &  &  &  &  &  &  \\
\midrule
1 & 18.09 ± 1.15 & 46.30 ± 3.71 & 107.92 ± 4.98 & 156.10 ± 6.73 & 6.38 ± 1.50 & 20.26 ± 3.04 & 35.50 ± 4.99 & 68.36 ± 7.90 \\
2 & 19.66 ± 2.30 & 55.33 ± 3.86 & 113.94 ± 5.17 & 167.13 ± 5.77 & 7.25 ± 0.97 & 24.13 ± 3.72 & 40.57 ± 3.00 & 78.99 ± 5.97 \\
3 & \textbf{20.71} ± \textbf{2.60} & \textbf{62.57} ± \textbf{6.77} & \textbf{124.81} ± \textbf{6.68} & \textbf{181.81} ± \textbf{11.80} & \textbf{7.73} ± \textbf{1.37} & \textbf{25.30} ± \textbf{2.96} & \textbf{47.17} ± \textbf{3.57} & \textbf{88.43} ± \textbf{5.68} \\
\bottomrule
\end{tabular}
}
\caption{Adding agents---with a shared policy and shared reward---improves performance and generalization on a maze level generation task, both on (fixed, $16\times 16$) maps seen during training, and on maps of different maximum sizes and/or with randomized rectangular shapes. Agents are allowed twice as many environment steps as there are tiles on the map. Values are averaged over models from 10 training run seeds, each evaluated over 50 episodes in each evaluation setting.
% \Eugenei{Has it already been defined at this point what it means for rand. map shape to be true or false?}
}
\label{tab:ma_n_agents_eval_randomize_map_shap_map_width}
\end{table*}

This paper investigates whether a multi-agent approach can improve results in PCGRL. % chiefly by making training more efficient. 
% It is plausible that there are other advantages to multi-agent PCGRL as well, such as in functional specialization of agents. These will be discussed, however the main objective of the current paper is increased training efficiency.
%END OF NEW JULIAN TEXT
% As a benchmark for reinforcement learning algorithms, video game level design is a complex and challenging task, more so than mere inverse solving.
% One component of the task we investigate here---creating a maze with maximum diameter, or longest-shortest traversible path between two tiles---is a non-trivial combinatorial optimization problem. 
% Accordingly, PCGRL is a long-horizon problem with relatively sparse reward, making it challenging to approach with RL techniques. 
Because PCGRL is a long-horizon problem involving rewards that depend on global, combinatorial features that may be hard to model, it is challenging to approach with RL techniques. 
Prior work has observed that decreasing the size of the observations that agents use, from observing the full map to solely observing a small local region, leads to empirical improvements in efficacy and generalization~\cite{earle2024scaling}. This finding suggests that although maze-like level construction does involve global reasoning, much of the task can be accomplished locally. This in turn suggests that the problem can be factorized without much loss into sub-problems. 
% Such sub-problems are only partially decoupled, they have hints of the global structure through the edges of the observed boundaries (TODO, very confusing sentence).

We hypothesize that decomposing PCGRL for game level construction into a decentralized problem, in which multiple agents seek to cooperatively build a level, may yield significant improvements in PCGRL capabilities\sam{maybe efficiency? I guess capabilities makes sense as combining generalization and efficiency}. Agents can efficiently learn to construct levels for the local region they are tasked with, while approaching global coherence by exchanging information via overlapping regions that they jointly observe.

This type of decomposition, if possible, is critical for efficient PCGRL. By converting the single-agent PCGRL problem into a multi-agent one, we achieve several key benefits. First, the horizon of the problem is potentially reduced, with each agent taking fewer actions before the entire level has been edited by the collective. Second, for vectorized simulators where the cost of stepping agent dynamics is low, it is often as fast to simulate one agent as it is to simulate many. Instead, for PCGRL environments, the step-time of the environment is heavily dominated by the cost of computing rewards at each step. 
% On average, for $N$ agents, we will have to take $T/N$ total steps where $T$ is the average horizon of the problem. 
By sharing reward amongst the agents at each step, we can reduce the cost of stepping the most expensive part of the environment.
Finally, the factorization of PCGRL tasks into local sub-problems may yield generalization benefits as has been observed in other work~\cite{earle2024scaling}. 

% Given that the predominant computational expense comes from reward function calculation, it makes sense to have the reward function evaluate the results of multiple actions. But having the same agents take multiple actions sequentially will mean a more sparse reward. Instead, we let multiple agents output actions in parallel, and calculate the reward after the entire batch of actions has been executed. This allows us to take more actions per reward function call, while preserving a dense reward scheme.

% In this work, we perform an empirical investigation of the efficacy of turning PCGRL into a multi-agent problem. We design a JAX-accelerated multi-agent learning loop, allowing us to generate X samples per second. 

In sum, our contributions are as follows:
\begin{enumerate}
    \item We frame video game level design as a multi-agent reinforcement learning problem.
    % \item We extend a popular framework---PCGRL---for level design via RL to support the multi-agent setting, in which the training loop and environment is parallelized and runs entirely on the GPU to allow for rapid experimentation.
    \item We extend the PCGRL framework for level design via RL to support a multi-agent setting, implemented in JAX. The training loop and environment are fully parallelized and run entirely on the GPU, enabling rapid experimentation.
    \item We investigate the effect on performance and generalization of the number of agents, number of episode steps, and reward-computation frequency, showing that additional agents provide performance, generalization and efficiency gains relative to single-agent PCGRL.
\end{enumerate}

\section{Related Work}

\subsection{Procedural Content Generation via Reinforcement Learning}
Procedural Content Generation via Machine Learning (PCGML, \citealt{summerville2018procedural}) is an active field of research that explores the capacity for AI systems to generate game content. Much research has been conducted in supervised and self-supervised PCGML methods for level generation.
% These levels are often represented as two-dimensional grids of discrete values---each value corresponding to a "tile" in the level.
% The simplest generative pipeline utilizes supervised learning. 
Assuming a dataset of game levels, models can be trained using supervised learning to generate similar game levels~\citep{snodgrass2016learning,summerville2016super,merino2023fivedollar}. Self-supervised methods such as Generative Adversarial Networks and Variational Autoencoders have also exhibited success in level generation~\citep{torrado2019bootstrapping, toadgan, sarkar2021generatingblendinggamelevels}. Recent research has even explored the capacity of Large Language Models as procedural content generators \cite{Todd_2023, sudhakaran2023mariogpt,merino2024making,earle2025dreamgarden}.
% A main drawback of these methods is the data requirement. To generate content for a game, we first need a dataset of the content we want to generate. This begs the question: \textit{Why do we need a generator if we already have a large dataset of existing content?}

% However, designers for new games, for which no data is available, might also benefit from controllable level generators.
% And in some domains where function reigns over content, 

\begin{figure*}[t]
\centering

% \begin{subfigure}{\linewidth} 
% \begin{subfigure}{.24\linewidth} 
% \includegraphics[width=\linewidth]{imgs/2_agents_hemera/enjoy_10820-0.png}
% \caption{$t=0$}
% \end{subfigure}
% \begin{subfigure}{.24\linewidth} 
% \includegraphics[width=\linewidth]{imgs/2_agents_hemera/enjoy_10820-114.png}
% \caption{$t=114$}
% \end{subfigure}
% \begin{subfigure}{.24\linewidth} 
% \includegraphics[width=\linewidth]{imgs/2_agents_hemera/enjoy_10820-227.png}
% \caption{$t=227$}
% \end{subfigure}
% \begin{subfigure}{.24\linewidth} 
% \includegraphics[width=\linewidth]{imgs/2_agents_hemera/enjoy_10820-1440.png}
% \caption{$t=1440$}
% \end{subfigure}
% \caption{2 agents}
% \end{subfigure}

\begin{subfigure}{.85\linewidth} 
\begin{subfigure}{.24\linewidth} 
\includegraphics[width=\linewidth]{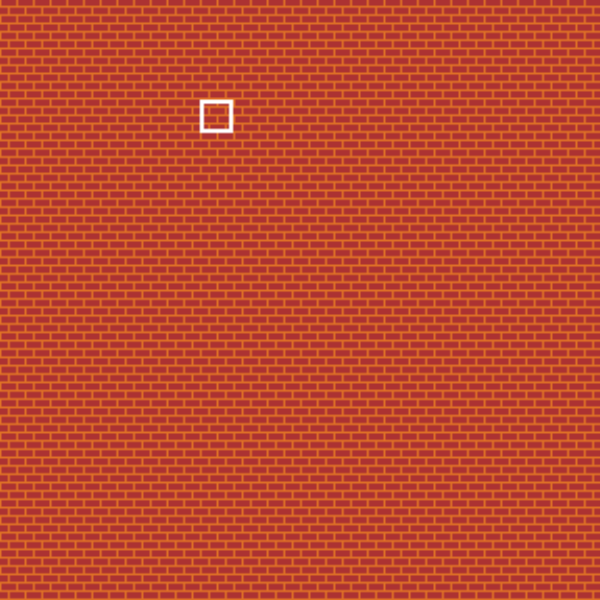}
\caption*{$t=0$}
\end{subfigure}
\hfill
\begin{subfigure}{.24\linewidth} 
\includegraphics[width=\linewidth]{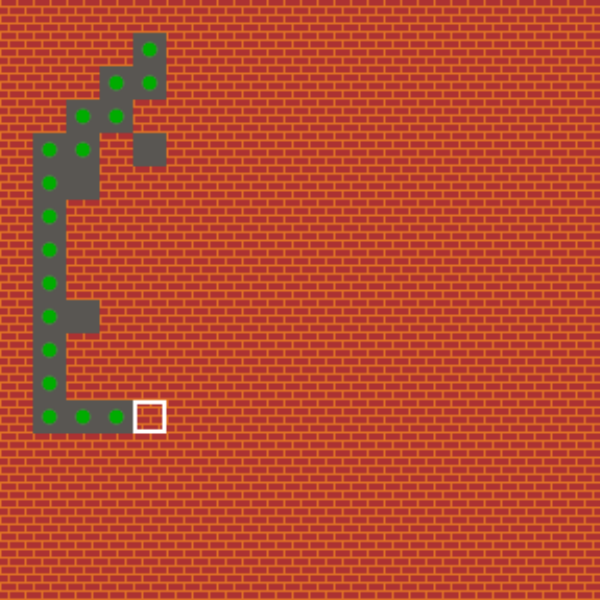}
\caption*{$t=188$}
\end{subfigure}
\hfill
\begin{subfigure}{.24\linewidth} 
\includegraphics[width=\linewidth]{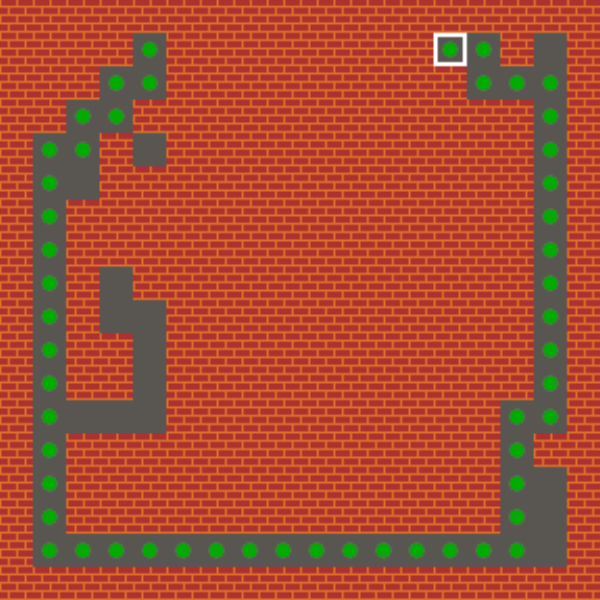}
\caption*{$t=643$}
\end{subfigure}
\hfill
\begin{subfigure}{.24\linewidth} 
\includegraphics[width=\linewidth]{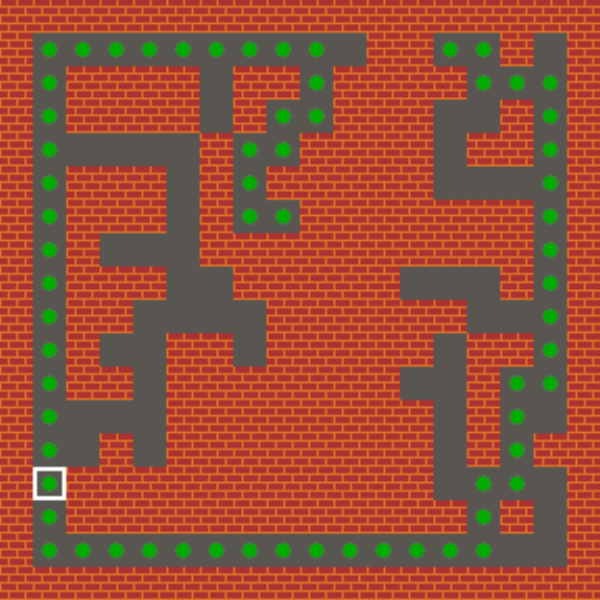}
\caption*{$t=1835$}
\end{subfigure}
\caption{1 agent}
\end{subfigure}

\begin{subfigure}{.85\linewidth} 
\begin{subfigure}{.24\linewidth} 
\includegraphics[width=\linewidth]{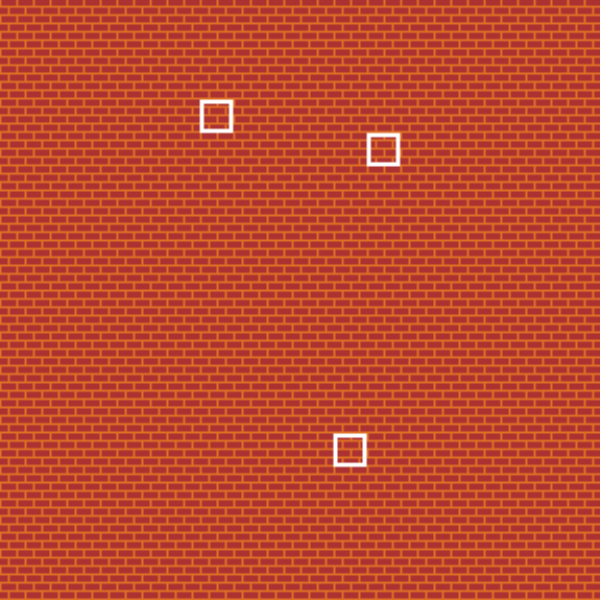}
\caption*{$t=0$}
\end{subfigure}
\hfill
\begin{subfigure}{.24\linewidth} 
\includegraphics[width=\linewidth]{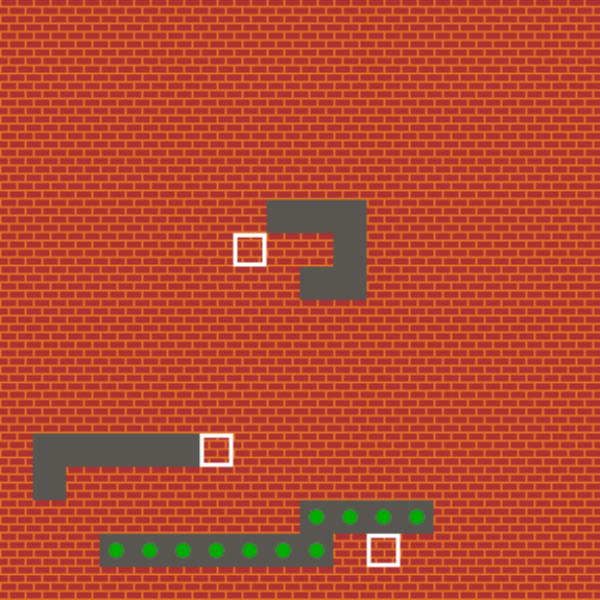}
\caption*{$t=65$}
\end{subfigure}
\hfill
\begin{subfigure}{.24\linewidth} 
\includegraphics[width=\linewidth]{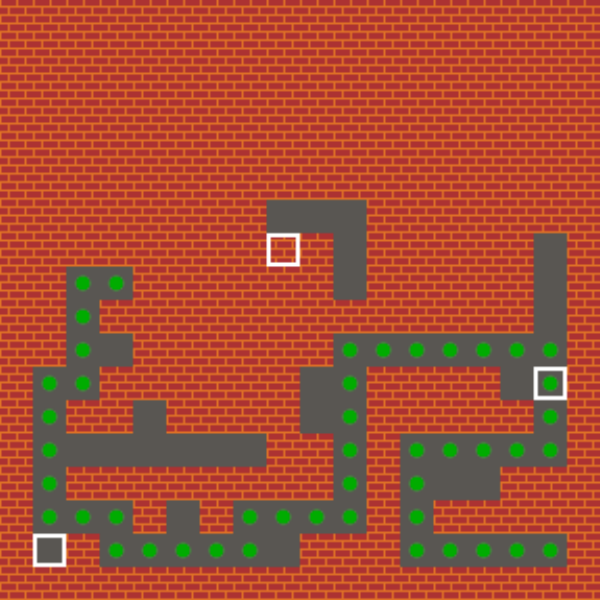}
\caption*{$t=229$}
\end{subfigure}
\hfill
\begin{subfigure}{.24\linewidth} 
\includegraphics[width=\linewidth]{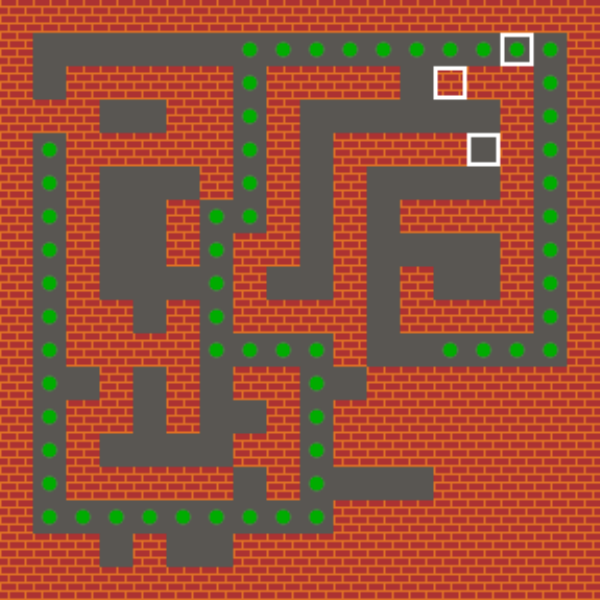}
\caption*{$t=1508$}
\end{subfigure}
\caption{3 agents}
\end{subfigure}

\caption{Episode rollouts with variable numbers of agents. Having multiple agents tends to allow for better map coverage and more complex and modular design patterns. For the sake of illustration, levels are initialized full of wall tiles. $t$ represents the timestep of a given frame. \sam{I'm not sure this is convincing that the second one has better board coverage than the first. Like, maybe a little?}}
\end{figure*}

Procedural Content Generation via Reinforcement Learning (PCGRL) is a subset of PCGML that prioritizes the functional characteristics of generated levels over their aesthetics.
In a sense, it represents an inversion of the typical RL paradigm: rather than training an agent to \textit{play} game levels, PCGRL trains an agent to \textit{create} them. 
% This shift in domain necessitates a re-framing of the creative process---rather than evaluating a level as a whole, based on high-level aesthetic or functional criteria, we view it as 
To this end, level design is itself cast as a kind of game---an iterative process in which agents make local edits in order to maximize functional metrics over a long time horizon.
% (a strictly more complex task).
Single-agent PCGRL was first explored by \citet{pcgrl}, where they applied the method to generate 2D tile-based binary mazes, Zelda-like dungeons and Sokoban levels.

% In PCGRL, level generation is formulated as a Markov Decision Process (MDP). At each step, the agent gets an observation and reward, and decides on an action. Rather than traditional PCGML approaches, which generate an entire level in a single forward pass, PCGRL frames level generation as an iterative task where levels are modified one tile at a time. Each episode begins with one $(N, N)$ array of tiles, which are randomly initialized at the start from an existing tile set. At each time step, the agent is allowed to change a single tile, and receives a reward based on how that change brings the agent closer to the desired goal state \citep{pcgrl}.
% In PCGRL, level generation is modeled as a Markov Decision Process (MDP).
In PCGRL, level generation is modeled as a sequential decision making process.
At each step, the agent receives an observation, a reward, and selects an action. Unlike traditional PCGML methods that generate entire levels in a single pass, PCGRL generates levels by modifying a single tile at a time. Each episode starts with an $N^2$ array of tiles randomly initialized from a predefined, discrete tile set. The agent changes a single tile per time-step and earns a reward based on how the change affects certain global functional characterstics of the level relative to some target values (e.g. a desired path-length or number of connected traversible regions). In follow-up work, the PCGRL framework was extended to support variable functional targets, with agents learning to produce diverse levels in response to conditional inputs \citet{controllable}. \citet{jiang2022learning} apply PCGRL to 3D mini-games in MineCraft, and \citep{earle2024scaling} further probe the approach's scalability by training generators to work around ``frozen'' pivotal tiles and evaluating generalization on out-of-distribution map shapes.

Notably, PCGRL does not require any training data; computational costs are largely front-loaded into the training process. Once a generator has been trained, inference is cheap, a critical component in enabling real-time content generation. This stands in contrast to search-based PCG~\cite{togelius2011search}, which has seen limited adoption in run-time content generation in games largely because of its high latency in combinatorially complex search spaces.

% well-explored supervised learning approaches to procedural content generation, where the computational costs of inference at scale can be similarly prohibitive to the computational costs of training.

PCGRL has two major intended applications.
One is to produce efficient generators for environments in which to train robust and generally-capable embodied agents.
Along these lines, PCGRL has influenced the level generation approaches in Unsupervised Environment Design~\cite{dennis2020emergent,jiang2021prioritized,parker2022evolving}. \sam{cite PAIRED... is this fair? Anything else?}
\Eugene{I think you could cite probably anything and everything in minimax https://github.com/facebookresearch/minimax}
The other application is to produce effective generators for use in co-creative pipelines alongside human designers.
Much prior work has examined the use of real-time level-generators as co-creative tools, using methods such as interactive evolution~\cite{schrum2020interactive}, quality diversity evolutionary search~\cite{charity2020baba}, and supervised learning~\cite{guzdial2019friend}.
\citet{delarosa2021mixed} explore the use of PCGRL agents as design assistants with \textit{RL Brush}, an online tool for Sokoban level design in which users can select between suggested edits from a diverse set of trained models, make manual edits, and playtest partial levels.\sam{todo. More co-creative cites?}
\TheTMan{Tim poopery: maybe move this elsewhere or scrap? idk brain blast}

\subsection{Multi-Agent Reinforcement Learning}
Multi-agent RL has been applied both to real-world control problems, such as coordination of drone swarms \citep{droneswarms} and autonomous vehicle fleets \citep{driving}, and virtual ones, such as the control of multi-unit teams in the real-time strategy video game \textit{StarCraft}~\cite{ellis2023smacv2}.
While the objectives of agents in the above domains are shared, they may also be competitive or mixed~\citep{marloverview}.
In Artificial Life-like simulations like NeuralMMO, for example, per-agent rewards lead to a variety of emergent and variably selfish strategies among agents in terms of combat, foraging and niche-finding~\cite{suarez2023neural}.
Assuming that designers have a fixed idea of desirable level features, PCGRL is naturally suited to a cooperative framing, though future work could explore whether novel design patterns emerge when synthetic designers with orthogonal objectives are forced to collaborate.

\citet{mappo} investigate the effectiveness of PPO-based RL methods in various cooperative multi-agent settings. They find that PPO-based multi-agent algorithms are competitive with off-policy RL methods such as MADDPG and QMIX. Given this, we find it sufficient to apply Multi-agent PPO (MAPPO) to our PCGRL environment (and leave comparison against other algorithms to future work).

% We explore how Multi-agent reinforcement learning can be incorporated into PCGRL training. By incorporating multiple generator agents into the training process, we hope to exploit the potential benefits of MARL while investigating interesting emergent behaviors in a level generation domain.

JAX~\cite{jax2018github} is a library that allows for a broad array of numpy-type operations \cite{harris2020array} to be just-in-time compiled and run on the GPU, taking advantage of array parallelism.
It has been used to accelerate RL environments in both single-agent~\cite{coward2024jaxued}\sam{more cites}, and multi-agent settings~\cite{rutherford2024jaxmarl}.
PCGRL has been ported to JAX~\cite{earle2024scaling}, allowing for a $\approx17\times$ speedup in training time relative to the numpy version \cite{pcgrl}.

 The idea of multi-agent or co-creative PCG has been explored in a few different domains, from using hand-written agents for city planning \cite{city-building}, to co-creative design of perfume bottles \cite{cocreative}, to exploring a variety of generative models, ranging from locally to globally focused, for co-creative PCGML for level designs in Super Mario Bros. \cite{perfume_design}. This work is the first, to our knowledge, that treats PCG as a multi-agent RL problem.

\section{Methods}

% We extend this existing work by first re-implementing multi-agent PCGRL in the updated JAX framework~\cite{jax2018github}. Our work focuses on multi-agent dynamics in the \textsl{binary} domain from \citet{pcgrl} primarily. 

We extend previous work by adding multi-agent support to the JAX version of the PCGRL framework, focusing on the \textsl{binary} and \textsl{dungeon} domains using the \textsl{turtle} representation from \cite{pcgrl}.

% In this domain, the agents' are able to place ``wall'' tiles, which are non-traversible, and ``air'' tiles which are traversible. The current goal is to generate a maze with maximum diameter (the longest shortest path between two air tiles), while having only 1 4-neighbor connected component of air tiles. The diameter of the maze is approximated using Dijkstra's algorithm, and connected components are counted using a flood fill algorithm \citep{pcgrl}.

In the \textsl{binary} domain, `turtle' agents can place impassable `wall' tiles or traversible `air' tiles, or move to adjacent positions. At the start of each episode, agents are randomly positioned on the map, which is initialized uniformly with a mixture of wall and air tiles.
The objective of the generator agents is to construct a maze with maximal diameter (defined as the longest shortest path between any two air tiles) while maintaining only one connected traversible component. Diameter is approximated using two passes of Dijkstra's algorithm. Agents are rewarded based on how much their actions reduce the distance between the values of these metrics in the current state and their target values (viz. the maximum possible diameter, and $1$ connected component).

In the \textsl{dungeon} domain, additional constraints are introduced relative to the binary maze domain. Each level includes a player, a key, a door, and several enemies. In a dungeon, the implied goal of the player would be to collect the key, reach the door, and avoid enemies. So for a level to be considered playable, it must contain exactly one player, one key, and one door.
% The path finding algorithm remains the same as in the binary domain. 
Three shortest paths are computed: from the player to the key, from the key to the door, and from the player to the nearest enemy. These paths are computed only after the correct number of game elements is present. The nearest enemy must be at least three tiles away from the player, and the player's two-part path to unlocking the door should be maximal. 
The dungeon domain introduces more complex gameplay constraints compared to the binary domain, allowing us to evaluate the agents in scenarios that more closely resemble real-world use cases with diverse requirements and constraints.

% During training, each episode ends after a fixed number of agent actions, preventing the level from changing too much. This is determined by either a change percentage, the percentage of tiles that the agent can alter, or a fixed number of time steps before the episode terminates. 
% To address potential mode collapse in the PCGRL environment, we limit episode length to less than a full board scan for each agent during training, forcing the agent to adapt to the random initial state and choose the most effective actions.

% We control the number of environment steps allowed in an episode relative to the number of steps that would be required in the single-agent case to scan the entire board (given an optimally space-filling trajectory from a turtle agent), or ``board scans''. (I.e. ``1 board scan''-many steps over a $16\times 16$-tile maps, in the case of a \textsl{turtle} representation, corresponds to $256\times2=512$ steps, allowing the agent to spend half of its actions moving along a trajectory covering the entire map, and the other half editing each tile along this trajectory).

To enable more meaningful comparisons across hyperparameter sweeps and experimental settings, we standardize the number of environment steps allowed in an episode using a unit we call a “board scan.” A board scan represents the number of steps a single agent would need to traverse the entire board using an optimally space-filling trajectory (e.g., following a \textsl{turtle} pattern). For instance, on a $16 \times 16$ map, one board scan corresponds to $256 \times 2 = 512$ steps, allocating half the steps for movement and the other half for editing each tile along the path.
% When this value of maximum allowed board scans is less than $1$, in the single agent case, the agent cannot edit all tiles on the map, preventing overfitting or mode collapse via memorization of a single optimal level, and instead forcing the agent to incorporate some of the initial noise into its design.
When this value of maximum allowed board scans is less than $1$, in the single agent case, the agent cannot edit all tiles on the map and is thereby forced to incorporate some of the initial noise into its design, which constraint we would expect to discourage mode collapse or memorization of a single optimal level layout.

% This forms the whole process into a finite horizon MDP, where the agent-environment loop can be understood as follows: At each step the agents observe the current state of the environment ($S_t$). This state is fed into the agent's neural network, which provides an action ($a_t$). Based on this action, the state changes ($S_{t+1}$) and the agent receives a reward ($r_{t+1}$) based on the difference between $S_t$ and $S_{t+1}$. This continues until the end of the episode.

% We focus on two agents for most tests, which can traverse the map by moving 1 tile in any of the four cardinal directions at each step (this is the ``turtle'' representation from the prior work). 

\begin{table*}[ht]
\centering
\adjustbox{max width=\textwidth}{%
% \scriptsize
\begin{tabular}{ccllllllll}
\toprule
 &  & \multicolumn{8}{c}{mean episode reward} \\
 & map shape & \multicolumn{4}{c}{fixed} & \multicolumn{4}{c}{random} \\
 & map width & 8 & 16 & 24 & 32 & 8 & 16 & 24 & 32 \\
\toprule
n. agents & max. board scans &  &  &  &  &  &  &  &  \\
\midrule
1 & 1.50 & \textbf{14.60 ± 0.63} & \textbf{42.49 ± 1.20} & \textbf{78.85 ± 1.83} & 120.65 ± 1.90 & 5.81 ± 0.44 & 11.77 ± 0.63 & 10.25 ± 0.53 & 16.53 ± 0.39 \\
2 & 0.75 & 14.13 ± 0.92 & 40.97 ± 1.63 & 72.47 ± 2.75 & \textbf{114.47 ± 3.38} & \textbf{7.47 ± 0.12} & \textbf{15.29 ± 0.36} & \textbf{22.25 ± 0.75} & 21.55 ± 1.34 \\
3 & 0.50 & 11.89 ± 0.54 & 36.24 ± 3.64 & 67.64 ± 4.35& 108.19 ± 2.85 & 6.53 ± 0.26 & 11.36 ± 0.19 & 20.95 ± 1.34& \textbf{34.80 ± 0.27} \\
\midrule
1 & 3.0 & \textbf{16.31 ± 0.43} & 44.68 ± 0.42 & 79.49 ± 2.42 & 113.19 ± 9.54 & 7.39 ± 0.11 & 13.47 ± 0.53 & 10.41 ± 0.50 & 16.00 ± 0.28 \\
 2 & 1.5 & 14.71 ± 0.07 & \textbf{46.23 ± 1.40} & \textbf{87.14} ± \textbf{2.18} & \textbf{130.92} ± \textbf{3.49} & 7.32 ± 0.33 & 14.17 ± 0.22 & 18.81 ± 0.78 & 30.29 ± 1.30 \\
3 & 1.0 & 15.56 ± 0.63 & 44.49 ± 1.99 & 83.94 ± 5.12 & 124.95 ± 10.47 & \textbf{8.66} ± \textbf{0.47} & \textbf{14.82 ± 0.41} & \textbf{25.55 ± 1.58} & \textbf{41.52} ± \textbf{2.81} \\
\bottomrule
\end{tabular}
} 
\caption{When controlling the number of environment steps (relative to the time it would take a single agent to scan through each tile on the map), so that the same number of actions are taken in situations with variable number of agents, settings with more agents generally attain comparable reward to those with fewer or a single agent, and do better in out-of-distribution settings where map shape is randomized at evaluation time. With more agents, reward computation---the bottleneck in the environment---occurs less frequently. Average of 5 training seeds. \sam{I think it's not clear how to tell which of these columns are in or out of distribution.}}
\label{tab:ma_board_scans_n_agents_eval_randomize_map_shap_map_width}
\end{table*}

\begin{figure}
    \centering
    \includegraphics[width=1\linewidth]{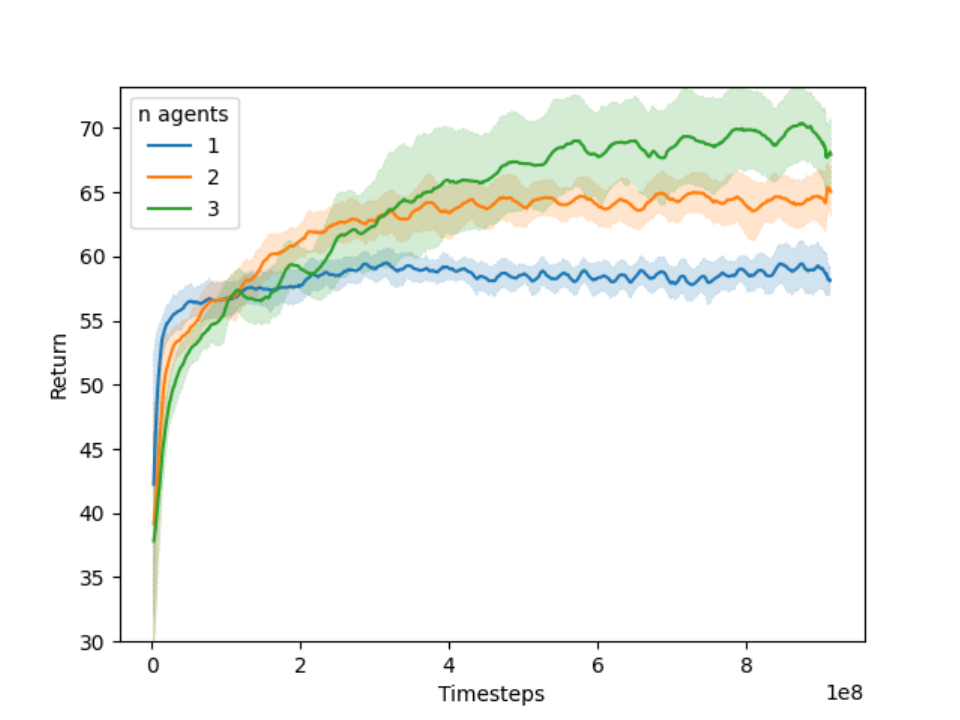}
    \caption{Episode returns by total number of environment steps during training, for varying numbers of agents. Results averaged over 10 training seeds.}
    \label{fig:ma_n_agents_metric_curves_mean}
\end{figure}

\begin{figure}[ht]
    \centering
    \includegraphics[width=0.8\linewidth]{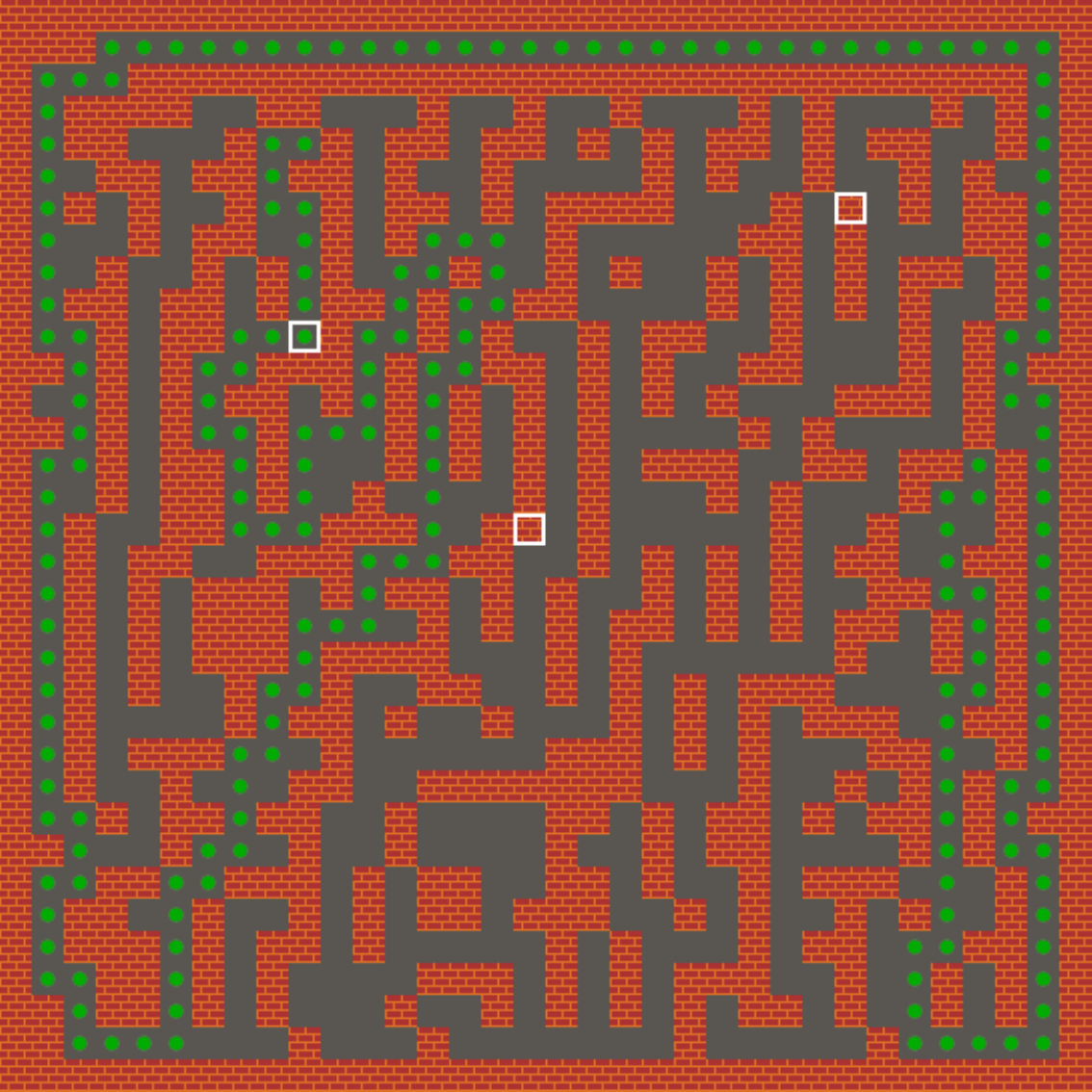}
    \caption{The result of three agents' collaboratively building a binary maze. The agents, trained on $16 \times 16$ maps, are evaluated on an out-of-distribution map of size $32 \times 32$.}
    \label{fig:best3agents}
\end{figure}

\begin{table*}
\centering
\adjustbox{max width=\textwidth}{%
% \scriptsize
\begin{tabular}{cccllllllll}
\toprule
& &  & \multicolumn{8}{c}{mean episode reward} \\
& & map shape & \multicolumn{4}{c}{fixed} & \multicolumn{4}{c}{random} \\
& & map width & 8 & 16 & 24 & 32 & 8 & 16 & 24 & 32 \\
\toprule
n. agents & max. board scans & reward freq. &  &  &  &  &  &  &  &  \\
\midrule
1 & 3.0 & 3 & \textbf{16.52} ± \textbf{0.21} & \textbf{46.33} ± \textbf{1.17} & 82.77 ± 0.90 & \textbf{125.41 ± 4.51} & 7.22 ± 0.25 & 13.65 ± 0.25 & 10.35 ± 0.50 & 15.69 ± 1.52 \\
3 & 1.0 & 1 &  15.56 ± 0.63 & 44.49 ± 1.99 & \textbf{83.94 ± 5.12} & 124.95 ± 10.47 & \textbf{8.66} ± \textbf{0.47} & \textbf{14.82 ± 0.41} & \textbf{25.55 ± 1.58} & \textbf{41.52} ± \textbf{2.81} \\
\midrule
1 & 3.0 & 2 & \textbf{16.36 ± 0.60} & \textbf{49.19} ± \textbf{0.99} & 80.81 ± 4.15 & 114.46 ± 17.52 & 6.93 ± 0.15 & 13.04 ± 0.35 & 10.37 ± 0.39 & 14.85 ± 0.10 \\
2 & 1.5 & 1 & 14.71 ± 0.07 & 46.23 ± 1.40 & \textbf{87.14} ± \textbf{2.18} & \textbf{130.92} ± \textbf{3.49} & \textbf{7.32 ± 0.33} & \textbf{14.17 ± 0.22} & \textbf{18.81 ± 0.78} & \textbf{30.29 ± 1.30} \\
\bottomrule
\end{tabular}
}
\caption{When adjusting reward computation frequency and number of board scans in order to hold constant the number of total agent actions and reward computations per episode between single and multi-agent scenarios, multi-agent runs perform slightly worse in-distribution, but generalize better relative to their single agent counterparts. Average of 3 training seeds.}
\label{tab:ma_sparse_reward_n_agents_eval_randomize_map_shap_map_width}
\end{table*}

\subsection{Reward}
\Eugene{Do you say what the reward function is here? It's implied later but we should say it upfront.}
The step function effectively applies a batch of tile-edit actions---one per agent---to the level in sequence.
(Conflicts resulting from agents occupying the same tile are resolved according to agent ordering.)
% The reward is calculated after every agent action, and is divided by the number of agents. 
% The reward for each full iteration over every agent's action is the sum of all those intermediary rewards. (I.e., each agent receives identical reward after each collective step in the environment.)
We implement a shared reward, in which all agents receive a reward corresponding to the degree to which the cumulative edits made by the population of agents bring the level closer to the desired heuristics.
% The framework allows for the possibility of different per-agent rewards via recombination of relevant metrics (or the potential inclusion of new ones).

\subsection{Multi-Agent Environment}

% We implement multi-agent PCGRL by layering the single-agent environments over the number of desirtd agents. We define a new class, containing the metadata of all agents within a pytree structure whose step function maps from batchxes of actions rather than single actions to a new state. 

% The environment also gives batched observation across agents. This is implemented similarly as above, scanning over all agents' positions in the multi-agent setting.
Each agent observes a local patch of the map centered at its current position (padded with special ``border'' tiles when this patch extends beyond the edge of the map).
To enable coordination between agents, per-agent binary masks are concatenated channel-wise to the map, each with a $1$ on whatever cell the relevant agent is located.
% These observations are returned as a dictionary with the agents as the keys.

% We further included the implementation for the possibility of different rewards in the code, but because of the short timeframe, leave this for our future work.

We also introduce a new hyperparameter---\textit{reward frequency}---which allows the user to set the interval, in terms of number of environment steps, between each reward computation. 
This feature may be generally useful in that it allows more infrequent reward computation, where this reward computation is the bottleneck in the time complexity of the environment, being as it involves path-length computations with complexity $O(N^2)$, where $N$ is the maximum width of a given map.
In the present work, it allows us to investigate whether any changes in performance or generalization resulting from additional agents might be attributable merely to increased effective reward sparsity (as all agents act at once and are rewarded together) instead of the multi-agent dynamics themselves.

\section{Experiments}
\subsection{Training}
We use Multi-Agent Proximal Policy Optimization (MAPPO)~\citep{yu2022surprising} to train our agents' shared neural network. 
% We use fully connected RNNs for simplicity for the Actor and Critic networks, and flatten all observations as input to the networks.
We use fully connected RNNs for the Actor and Critic networks, and flatten all observations as input to the networks.
All networks are implemented using the Flax framework~\citep{flax2020github}. 
% This will allow us, in the future, to use the standard JAX vectorization functions to generate multiple networks for multiple agents.
We conduct hyper-parameter sweeps and run the experiments on an HPC cluster, with each experiment executes on a single node equipped with an RTX 8000 GPU. Each experiment runs for $10^9$ total environment timesteps.

\Eugene{How many seeds are the results in this table over? It should probably say in the caption.}

% Based on the findings in (cite pcgrljax) in the single-agent PCGRL scenario, we focus on the effects of observation size on generative agents. (list the exact params used here?)
% independent variable of our experiments as this was found to be the variable most indicative of performance in the single-agent case (in forthcoming work on the reimplementation of PCGRL in jax).

% \subsection{Reward}

% We implement two reward schemes---shared reward and individual reward. In the individual reward scheme, each agent receives reward resulting only from its own action. Conversely, in the shared reward scheme, both agents receive the cumulative reward resulting from both actions. We found that the difference between each agent getting the immediate reward for their action or the reward being summed over all agents' actions resulted in dramatically different performance. With individual rewards, agent performance quickly reached a plateau, far lower than the single-agent baseline. When using shared reward, performance closely approximated the single agent baseline scenario.

\subsection{Observation Space}

We experiment with global and local observation windows for agents. In all cases, observation sizes are kept equal between agents. At each time step, each agent either observes the entire map state (a global observation), or a subsection of it (local observation); i.e. an egocentric patch of size $(2N-1)\times (2N-1)$ or $M\times M$ with $M < 2N-1$, respectively.
%Due to time constraints, we only experiment with the previously identified best local observation size of $$(8, 8)$$, versus global observations.
Any observation outside of the map is padded with a ``border'' tile value only used for padding.
% This is done both for the efficient JAX implementation and to allow agents to learn about the boundaries of the map.
\Z{add agent can observe each other}

The values of observation windows that we experiment over are $[3, 16, 31]$ where $31$ corresponds to a full observation of the map at any point, regardless of the agent's position within it.

% We use a map size of $16 \times 16$ tiles for all experiments as this allows for direct comparisons to the previous work while being large enough for complex behaviors and small enough for efficient training.
As in prior work on PCGRL, we train on maps of $16\times 16$ tiles, allowing for comparisons to prior single-agent approaches.
Generally speaking, this size is large enough to allow for complex and varied design strategies while remaining small enough to allow for rapid experimentation over broad sets of hyperparameters.

\subsection{Evaluation}

To assess the in-distribution performance of our agents, we evaluate them $50$ times, using different random evaluation seeds, on randomly-initialized $16\times 16$ maps, and record the average of cumulative episode reward.
To assess their out-of-distribution performance, following \cite{earle2024scaling}, we evaluate our agents (again with 50 evaluation seeds) using out-of-distribution square and rectangular map sizes. We use square maps of width $8, 16, 24$, and $32$.
When optionally randomizing map shapes, given one of these widths as the maximum size of the square board, we uniform-randomly sample the width and height of a rectangular map from within these values.
This evaluation penalizes agents that overfit by learning one or a handful of globally optimal maps on a given map size, as these ``memorized" maps will be either be sub-optimal or impossible to recreate on other map shapes and sizes.
In our experiments, each set of hyperparameters is repeated over a number of distinct training seeds.
We average results from the evaluation process above applied to models from different training seeds, and report standard deviations in tables (and average and report standard deviations over training metrics from these training seeds in training plots).
\Z{make this more clear}

\section{Results}

\subsection{Maze Domain}

The primary goal of our experiments is to investigate the effect of the number of agents on the performance and generalization ability of agents in a level-generation task.

\Eugene{The text here is not clear that in each case, the horizon is divided by the number of agents. Is that the case? I.e. it's not the case that in the 3 agent case, they also get to take 3x as many actions? If that's not the case, and they are taking more actions, you have to appeal to the fact that due to JAX the wallclock time is probably the same.}

% \subsection{}
In Table~\ref{tab:ma_n_agents_eval_randomize_map_shap_map_width}, we examine the effect of varying the number of agents, fixing the number of episode steps to be $2 \times$ the number of tiles on the map, and the agent observation windows to be $3 \times 3$ egocentric patches.
Note that in this experiment, we fix the number of \textit{environment steps}, and in effect the number of reward computations---the most expensive aspect of our environment due to the requisite path-length computations.
As a result, settings with more agents are able to make more editing passes on more of the overall map \textit{if} they act collaboratively, and do not merely repeat or undo one another's edits.
The results show that scenarios with more agents indeed perform and generalize better.
% The addition of each agent, up to $3$ agents, roughly doubles the cumulative episode reward in nearly all cases.
% For any number of agents, reward increases with board size, and---holding maximum board width constant---is lesser in evaluation settings in which map shapes are randomized within these maximum widths.
The performance gains resulting from additional agents can also be seen in the curves reflecting average reward over the course of training, shown in \ref{fig:ma_n_agents_metric_curves_mean}.

In Table~\ref{tab:ma_board_scans_n_agents_eval_randomize_map_shap_map_width}, we adjust the number of environment steps (relative to the total number of tiles in the map, or the maximum number of \textit{board scans} that could be achieved by a single agent given this many steps), so as to hold constant the number of individual tile edits allowed in cases with variable numbers of agents.
In effect, scenarios involving more agents are no longer capable of making more editing passes on a larger section of the map relative to single-agent scenarios, as they would have been in Table~\ref{tab:ma_n_agents_eval_randomize_map_shap_map_width}. Thus, any gains resulting from additional agents cannot be attributed to their ability to collectively make more edits to the map over the course of an episode.
In this scenario, some of the benefit of multiple agents is reduced relative to Table~\ref{tab:ma_board_scans_eval_randomize_map_shap_map_width}: multi-agent performance and generalization only approaches or is roughly comparable to that of a single agent on small maps with fixed shapes.
But significant improvements in generalization still hold in the most complex scenarios, on large maps with random shapes.
% In particular, adding a second agent while halving the number of allowed steps (from the 1.5 to 0.75 board scans) improves generalization on random map shapes.
% However, adding a third agent while further reducing the number of episode steps from 0.75 to 0.5 does not improve generalization beyond the 2-agent case, though it still strongly outperforms the single-agent case.

In Table~\ref{tab:ma_sparse_reward_n_agents_eval_randomize_map_shap_map_width}, we adjust the number of environment steps as well as the reward frequency---the number of steps after which we compute updated metrics and reward---so as to hold constant both the number of individual tile edits, \textit{and} the total number of reward computations, taken over the course of an episode.
In this case, any performance or generalization gains resulting from additional agents can be attributed neither to an increase in allowed map edits, nor to the sparsity of reward computations relative to agent edits.
Here, similar to Table~\ref{tab:ma_board_scans_n_agents_eval_randomize_map_shap_map_width}, we note that the quality of generated maps is similar in-distribution and on fixed-shape maps, but again improves significantly on large, randomly-shaped maps given more agents.
% Math out the number of total actions and reward computations and show that each pair of rows are equivalent?
% In particular, this is true when comparing 1 agent, with 3 board scans, and reward computations every 3 steps; 

\subsection{Dungeon Domain}

In Table~\ref{tab:ma_board_scans_dungeon_obs_3_eval_map_width_randomize_map_shap}, we replicate our initial multi-agent result (Table~\ref{tab:ma_n_agents_eval_randomize_map_shap_map_width}) in the dungeon domain.
Here, the generator must construct lengthy paths between a player and a key, a key and a door, and a player and a nearest enemy, and arrange for correct numbers of each of these special tile types. In this more complex domain, the advantages of multi-agent collaboration hold.

\subsection{Supplementary Results}

In Table~\ref{tab:ma_sparse_reward_eval_randomize_map_shap_map_width}, we vary reward frequency alone (in the single-agent case, while fixing the number of steps to be equivalent to one scan of the board, and setting a $3\times 3$ observation window), and find no significant differences in perforamce or generalization.

To validate our choice of local observations in the above experiments, in Table~\ref{tab:ma_obs_size} we investigate the effect of observation size in a multi-agent setting, in which 3 agents are allowed to take as many steps in the environment as equate to $1\times$ the number of tiles on the board.
We find that smaller, more local observations perform best, with $3\times 3$-tile local observations lead to the best performance and generalization, followed by $16\times 16$ and $32\times 32$ observations.
% The benefit of these local observations can also be observed in the reward curves over training steps in \ref{fig:ma_obs_size_metric_curves_mean}.

In Table~\ref{tab:ma_board_scans_eval_randomize_map_shap_map_width}, we conduct a more extensive sweep over number of steps (``board scans''), number of agents, and observation sizes.
We can see here that local observations are generally either similarly advantageous or comparable in the 2- and single-agent cases, and that the single-agent case is never dominant in any of the hyperparameter combinations considered.

\section{Discussion}

The fact that additional agents results in increased performance in all evaluation settings (including out-of-distribution map shapes and sizes) is strong evidence of the utility of training reinforcement learned level generators in the multi-agent setting.
Arguably, these results are somewhat surprising given the additional complexity involved in multi-agent control. In single-agent PCGRL, the next state/observation/reward can be deduced deterministically given the current state and an agent's action. Such a prediction becomes noisy in the multi-agent case, given the stochasticity in the actions of other agents, which is not observed until the following timestep.

Intuitively, when the number of environment steps is held constant, allowing for more agent actions in multi-agent cases, adding agents provides a clear advantage in terms of the amount of map edits and iteration afforded (Tables \ref{tab:ma_n_agents_eval_randomize_map_shap_map_width}, \ref{tab:ma_board_scans_dungeon_obs_3_eval_map_width_randomize_map_shap}).
And because the environment is GPU-accelerated, with agent actions executed in parallel, this increased coverage of the map comes at negligible cost in terms of wall-clock time.
% (especially when the number of steps is less than the amount required by a single agent to edit each tile).
But the fact that multi-agent scenarios outperform single-agent ones even when varying the number of environment steps to account for the discrepancy in total number of allowed actions shows that the performance gain cannot be attributed simply to an increased \textit{number} of possible actions.

Even in the case where a single agent can take as many actions as would be taken by multiple agents working together (Table \ref{tab:ma_board_scans_n_agents_eval_randomize_map_shap_map_width}), the latter still have a unique advantage in terms of spatial flexibility.
Supposing we allow enough actions for either set of agents to edit each tile in the board once, then a single agent must chart a course over the map using a single space-filling curve, whereas multiple agents may combine several smaller such curves to cover the map, allowing for a larger set of possible edit sequences to be enacted upon the map.
% \sam{Kind of a weird argument. I'm curious if there are more possible combinations of two half-sized space-filling curves relative to one large one. And is this necessarily a good thing?}
It may also be the case that by learning to adapt to each others' potentially noisy actions, agents learn more robust policies.\sam{I don't buy this tbh}

Generally, multi-agent scenarios increase training and run-time efficiency because they require fewer reward computations per agent action.
One might expect that such reward sparsity should, if anything, make the task more difficult, forcing agents to make longer time horizon associations between actions and reward.
But to be sure that the improved performance and efficiency gains of multi-agent scenarios are not attributable to this reward sparsity, we allow reward computation frequency to be adjusted (Table \ref{tab:ma_sparse_reward_n_agents_eval_randomize_map_shap_map_width}). Comparing multi-agent vs. single-agent settings in which the total number of actions and number of actions between each reward computation are held constant, we find that the generalization gains of multi-agent settings are maintained.

\begin{figure*}
\centering
\begin{subfigure}{1\linewidth} 
\begin{subfigure}{.24\linewidth} 
\includegraphics[width=\linewidth]{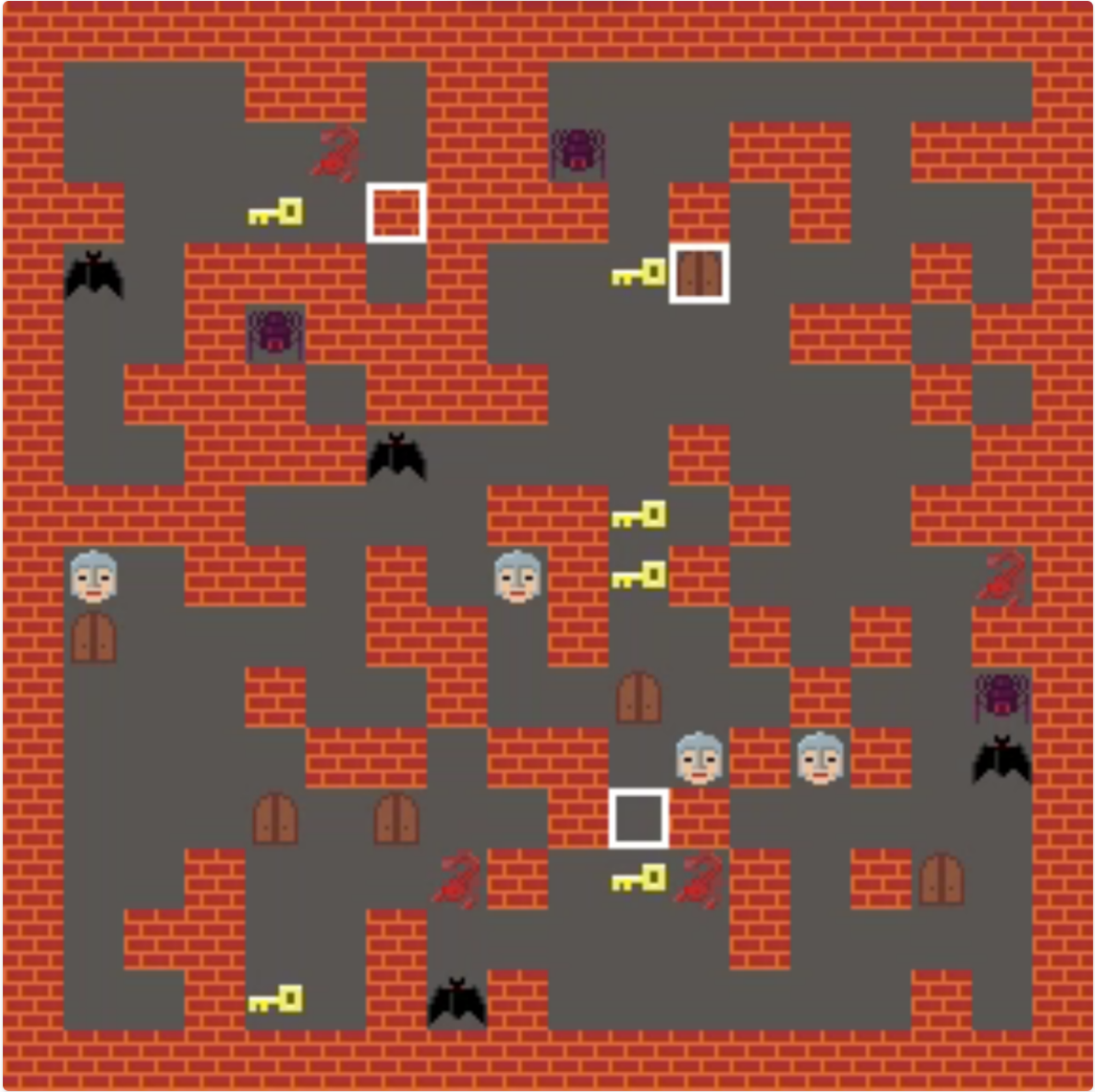}
\end{subfigure}
\begin{subfigure}{.24\linewidth} 
\includegraphics[width=\linewidth]{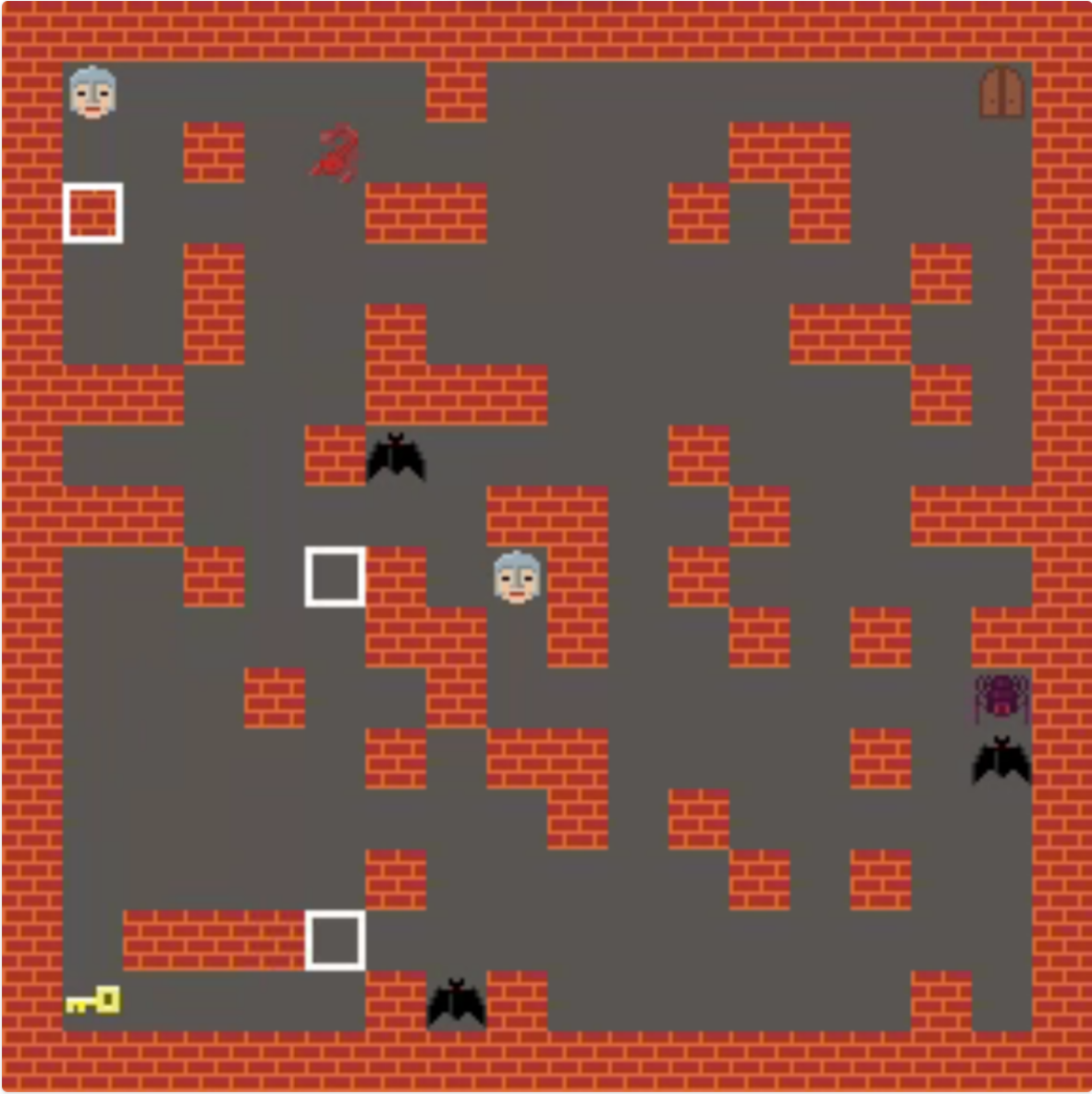}
\end{subfigure}
\begin{subfigure}{.24\linewidth} 
\includegraphics[width=\linewidth]{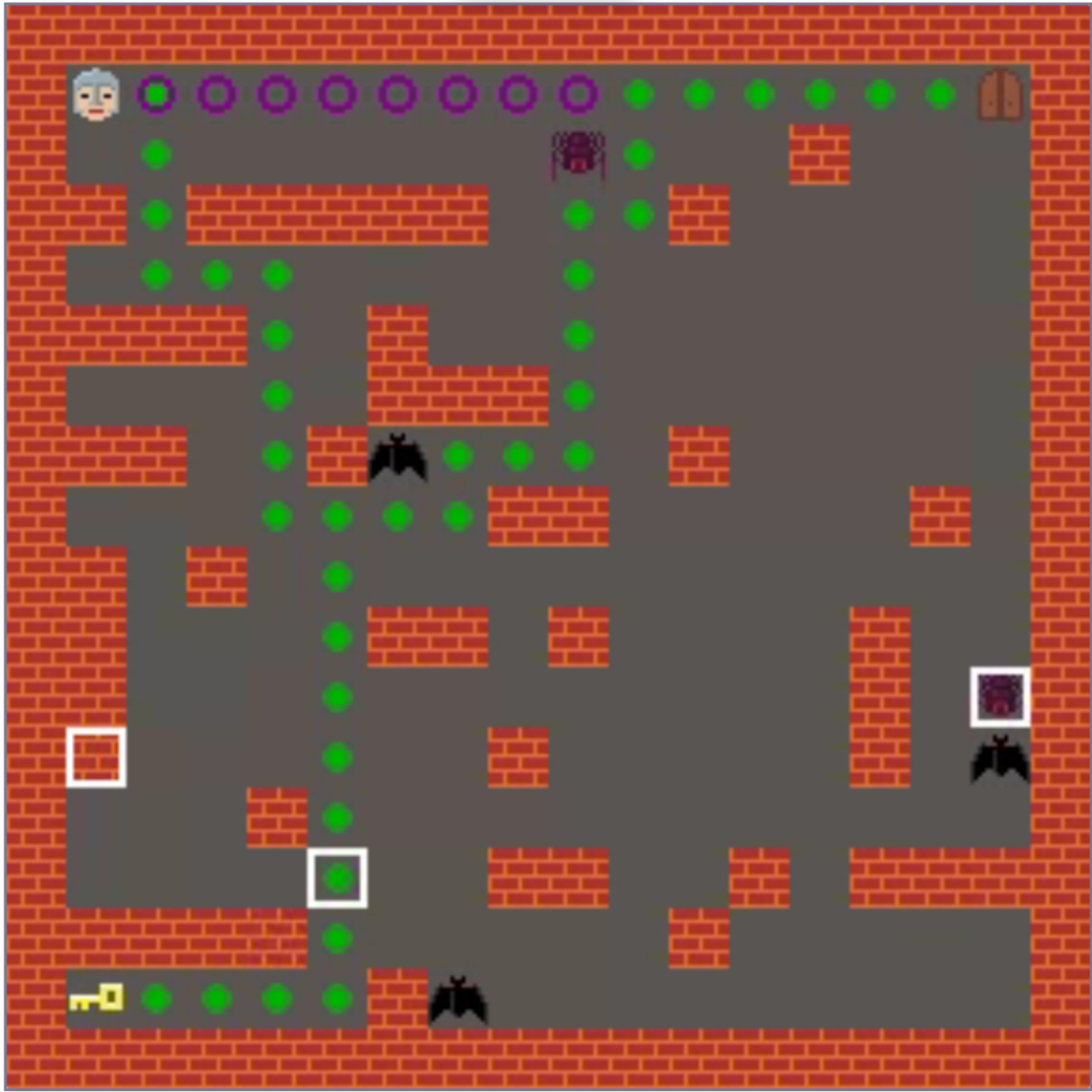}
\end{subfigure}
\begin{subfigure}{.24\linewidth} 
\includegraphics[width=\linewidth]{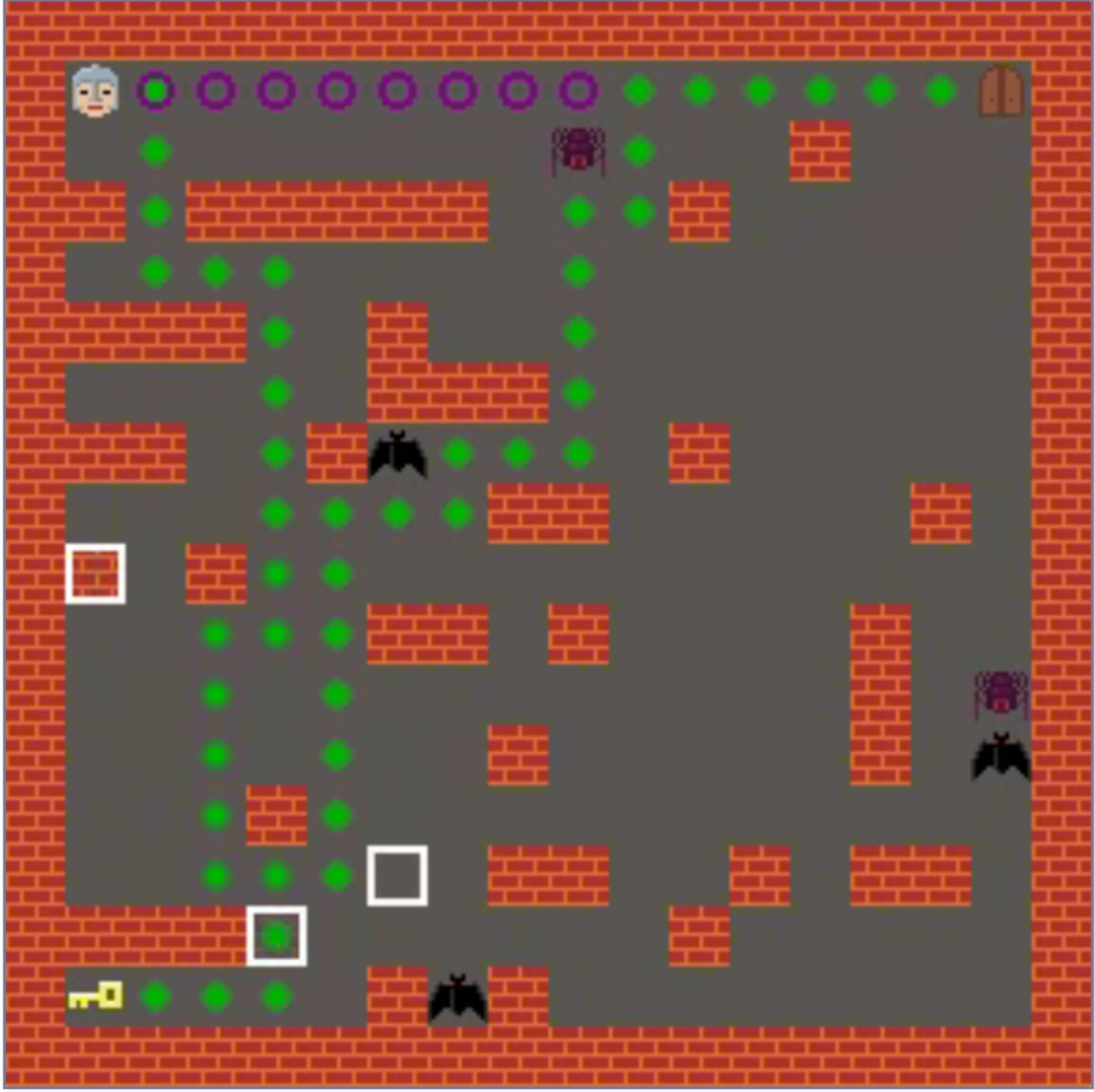}
\end{subfigure}
\end{subfigure}
\caption{Agents generate a level in the \textit{dungeon} domain.}
\end{figure*}

\begin{table*}
\centering
\adjustbox{max width=\textwidth}{%
% \scriptsize
\begin{tabular}{cllllllll}
\toprule
 & \multicolumn{8}{c}{mean ep reward} \\
map width & \multicolumn{2}{c}{8} & \multicolumn{2}{c}{16} & \multicolumn{2}{c}{24} & \multicolumn{2}{c}{32} \\
map shape & fixed & random & fixed & random & fixed & random & fixed & random \\
\toprule
n. agents &  &  &  &  &  &  &  &  \\
\midrule
1 & 11.96 ± 0.82 & 5.53 ± 0.50 & 66.43 ± 3.35 & 14.81 ± 0.95 & 129.77 ± 8.52 & 34.47 ± 2.30 & 213.78 ± 17.55 & 49.33 ± 4.00 \\
2 & 15.70 ± 0.59 & 7.68 ± 0.38 & 92.33 ± 3.47 & 21.86 ± 1.20 & 192.70 ± 10.37 & 51.06 ± 1.14 & 310.11 ± 29.95 & 78.06 ± 3.89 \\
3 & \textbf{16.99} ± \textbf{0.59} & \textbf{8.52} ± \textbf{0.36} & \textbf{99.91} ± \textbf{3.47} & \textbf{24.13} ± \textbf{0.95} & \textbf{214.17} ± \textbf{14.68} & \textbf{54.17} ± \textbf{3.04} & \textbf{364.47} ± \textbf{26.86} & \textbf{101.63} ± \textbf{6.09} \\
\bottomrule
\end{tabular}
}
\caption{Performance of multiple agents when designing levels in the ``dungeon'' domain. Here as elsewhere, an increase in the number of agents improves performance in and out of distribution while reducing computational cost by reducing the number of reward calculations. Average of 3 training seeds.}
\label{tab:ma_board_scans_dungeon_obs_3_eval_map_width_randomize_map_shap}
\end{table*}

Separately, we note that in the single-agent case, while reward sparsity does not significantly improve performance, it is also not detrimental, up to reward computation intervals of at least 10 agent actions (Table \ref{tab:ma_sparse_reward_eval_randomize_map_shap_map_width}).
This indicates that there is room to increase reward sparsity (in both single or multi-agent settings), making it possible to scale PCGRL to domains where reward computation is more expensive due to environment size or complexity.
%\footnote{It is worth noting that from an engineering standpoint, this would require some refactoring to avoid the evaluation of both ``branches'' of the JAX conditional determining whether to compute reward at each timestep.}  No that's not true lol, we know at what intervals this will be computed during compilation (i.e. we use a vanilla if statement to trigger these computations)

The fact that local observations handily outperform global ones is not surprising, given the empirical precedent revealed by prior work~\cite{earle2024scaling}, which suggests that a single feedforward agent is able to communicate globally relevant information to itself across time and space via stigmergy---i.e. implicit signals left by edits to the map.
It is nonetheless important to validate that this local approach still provides an advantage, as it is key to scaling learnable environment generation to larger levels where global observations are infeasible.
Because in this work we use a recurrent network, it is even less surprising that local observations work well, given that globally relevant facts can be stored in network memory.
Communication between agents, on the other hand, must still rely on the stigmergy of clues left in the environment.

\section{Limitations \& Future Work}

% When we implement multiple policies in the future we will also experiment with different training algorithms and training schemes.
Real-world application of our generative agents will involve levels that are larger, involving more diverse mechanics and a broader array of assets. 
Often, such tasks are tackled by teams of human designers, comprising multiple specialists.
% For example,  one designer may focus on creating a purely functional layout to capture the ``flow'' of the level, creating potential bottlenecks or areas of interest in a first-person-shooter such as \textit{Counter Strike}; while a second designer may focus on conceptualizing the narrative theme or semantics of the game world, contextualizing the layout of rooms, corridors, staircases and the likes via their aesthetic appearance.

By framing level design as a multi-agent task, we lay the groundwork for capturing this kind of specialization in teams of automated generators, making large-scale automatic environment generation feasible.
In the present work, we observe that some degree of specialization emerges naturally: often, separate agents will stick to different regions of the map, making changes to the level that, when taken on their own, would be locally sub-optimal, but which, when combined with the work of other agents, can push the level toward some global optimum.

Future work could push specialization further.
One simple extension could apply the ``pinpoint''-tile mechanic of \cite{earle2024scaling} to any edits made by a chosen \textit{priority} agent, which would then be subsequently unchangeable by other agents.
This would enforce specialization and potentially lead to the more rapid emergence of specialized behaviors among agents by explicitly preventing them from overwriting one another's decisions.
This may also have the beneficial side-effect of rendering the non-priority agent more adaptable to the actions of a collaborator, working to increase level quality given certain low-level constraints, making for better co-creative assistants that would be more performant when working alongside human designers.

% \todo[inline]{hierarchical agents---different brush strokes for different neural blokes}
Another common division of labor in human design teams is a hierarchical distribution of design tasks, in which some designers operate at a higher level of abstraction---for example delineating regions, terrain types, or building locations---while other designers focus on lower-level decisions---e.g. placing individual props, laying out paths or defining specific terrain topologies.
Multi-agent level design again allows for the instantiation of such a hierarchical work flow.
One simple means of realizing this in the current setting could be to afford different agents different ``brush stroke'' sizes: e.g. allowing one agent to transform entire $3\times 3$ patches of tiles to one type or another, while another agent is able to edit only individual tiles.

We must note that training an RL agent is a time-consuming affair (with the JAX PCGRL framework still requiring several hours---down from days in the numpy version of the framework---to train capable generators), and requires human designers to carefully engineer heuristics that provide learnable signals for agents. If these heuristics are simple enough to be encoded via more classical approaches such as constraint satisfaction, or ample human-authored data is available for supervised learning, then such approaches may provide a faster route to the automatic generation of satisfactory levels.

\section{Conclusion}

We frame the problem of collaborative video game level design as a multi-agent reinforcement learning problem, showing how embodied agents with local observations can self-organize over noisy initial levels to iteratively improve them, resulting in globally coherent and functional content.
We open-source our code, which builds on an existing, parallelized implementation of the Procedural Content Generation Reinforcement Learning (PCGRL) framework, allowing for rapid multi-agent training on the GPU.
While we limit our experimentation to the relatively bare-bones ``binary'' and ``dungeon'' PCGRL domains, our method is applicable to any future grid-based domains thay may be added to the framework. It may also be readily combined with existing PCGRL features (such as controllable high-level metrics or frozen tile placements) without further modifications to the code.

Our experiments demonstrate increases in runtime efficiency, performance, and generalization afforded by framing level design as a multi-agent task.
We find that even when holding the total number of agent actions constant, distributing these actions among multiple agents leads to stronger generalization ability, particularly on large maps with variable rectangular shapes.
We hypothesize that agents in such settings learn robust, modular strategies which are better suited to generalizing to these out-of-distribution scenarios.

Level design is a novel and complex task, distinct from the far more popular game-playing paradigm, that provide a unique set of challenges to multi-agent learning algorithms.
The successful automation of such a task has implications not only to the games industry, but also for other domains involving the iterative generation of globally coherent and functional artifacts.
Using multi-agent learning opens the door to automating such content generation tasks in a more efficient and scalable manner.

By producing agents that are naturally predisposed for collaboration, we can more readily produce content generators suited for use alongside human designers.
Future work should further explore the specialized and modular qualities inherited by multi-agent content generators, either by quantifying emergent behavior of learned generators, or enforcing specialization by design; and validate the utility of these generators in user studies with domain experts.

\appendix

\begin{table*}
\centering
\adjustbox{max width=\textwidth}{%
% \scriptsize
\begin{tabular}{lllllllll}
\toprule
 & \multicolumn{8}{c}{mean episode reward} \\
map shape & \multicolumn{4}{c}{fixed} & \multicolumn{4}{c}{random} \\
map width & 8 & 16 & 24 & 32 & 8 & 16 & 24 & 32 \\
\toprule
reward freq. &  &  &  &  &  &  &  &  \\
\midrule
1 & 13.64 ± 0.08 & 38.85 ± 1.21 & 68.38 ± 2.19 & 107.83 ± 1.24 & 5.82 ± 0.07 & 8.65 ± 0.16 & 17.42 ± 0.35 & 18.53 ± 0.30 \\
2 & \textbf{13.97} ± \textbf{0.24} & 38.80 ± 2.01 & 67.63 ± 2.94 & 105.48 ± 2.44 & \textbf{6.37} ± \textbf{0.26} & \textbf{8.71} ± \textbf{0.92} & \textbf{17.50} ± \textbf{0.69} & 18.53 ± 1.20 \\
3 & 13.59 ± 0.38 & 37.15 ± 0.97 & 66.28 ± 2.67 & 107.05 ± 4.36 & 6.23 ± 0.32 & 8.35 ± 0.30 & 16.86 ± 0.69 & 18.93 ± 0.66 \\
5 & 13.07 ± 0.27 & 38.33 ± 1.24 & 66.78 ± 3.45 & 103.51 ± 2.59 & 6.05 ± 0.13 & 8.67 ± 0.28 & 17.17 ± 0.17 & 18.73 ± 0.25 \\
10 & 11.44 ± 0.57 & \textbf{39.59} ± \textbf{0.66} & \textbf{69.39} ± \textbf{1.14} & \textbf{109.82} ± \textbf{6.81} & 4.86 ± 0.26 & 8.57 ± 0.76 & 16.65 ± 0.35 & \textbf{19.25} ± \textbf{0.27} \\
\bottomrule
\end{tabular}
}
\caption{Reward computation---the bottleneck of PCGRL complexity---can be performed at least as infrequently as every $10$ steps without hurting performance and generalization. However, it does not lead any clear improvement (unlike an increase in the number of agents, which similarly alleviates the need for frequent reward computation). Results for agents given steps-per-episode equal to one single-agent board scan. Average of 3 training seeds.}
\label{tab:ma_sparse_reward_eval_randomize_map_shap_map_width}
\end{table*}

\begin{table*}
\centering
\adjustbox{max width=\textwidth}{%
% \scriptsize
\begin{tabular}{lllllllll}
\toprule
 & \multicolumn{8}{c}{mean episode reward} \\
map shape & \multicolumn{4}{c}{fixed} & \multicolumn{4}{c}{random} \\
map width & 8 & 16 & 24 & 32 & 8 & 16 & 24 & 32 \\
\toprule
obs. size &  &  &  &  &  &  &  &  \\
\midrule
3 & \textbf{46.68} ± \textbf{1.88} & \textbf{133.48} ± \textbf{5.96} & \textbf{251.82} ± \textbf{15.35} & \textbf{374.84} ± \textbf{31.41} & \textbf{25.98} ± \textbf{1.40} & \textbf{44.46} ± \textbf{1.24} & \textbf{76.64} ± \textbf{4.75} & \textbf{124.56} ± \textbf{8.42} \\
16 & 30.58 ± 2.98 & 104.44 ± 1.86 & 185.06 ± 20.51 & 290.38 ± 33.68 & 17.32 ± 1.99 & 32.02 ± 0.67 & 63.02 ± 3.25 & 97.56 ± 5.68 \\
31 & -1.30 ± 3.00 & 117.46 ± 3.49 & 166.12 ± 3.63 & 260.42 ± 6.81 & 2.20 ± 2.27 & 14.46 ± 2.53 & 48.24 ± 4.55 & 85.36 ± 1.58 \\
\bottomrule
\end{tabular}
}
\caption{In a multi-agent setting---with 3 agents editing the map collaboratively---restricting agents' observation windows to local, $3\times3$ patches improves performance and generalization. Average of 3 training seeds.}
\label{tab:ma_obs_size}
\end{table*}

\begin{table*}
\centering
\adjustbox{max width=\textwidth}{%
% \scriptsize
\begin{tabular}{lllllllllll}
\toprule
 &  &  & \multicolumn{8}{c}{mean episode reward} \\
 &  & map shape & \multicolumn{4}{c}{fixed} & \multicolumn{4}{c}{random} \\
 &  & map width & 8 & 16 & 24 & 32 & 8 & 16 & 24 & 32 \\
\toprule
max. board scans & n agents & obs size hid dims &  &  &  &  &  &  &  &  \\
\midrule
\multirow[t]{9}{*}{0.75} & \multirow[t]{3}{*}{1} & 3 & 12.52 ± 0.47 & 41.39 ± 2.38 & 67.25 ± 2.50 & 96.63 ± 1.40 & 4.81 ± 0.14 & 11.77 ± 0.29 & 12.29 ± 0.88 & 14.56 ± 0.76 \\
 &  & 16 & 10.50 ± 0.48 & 45.24 ± 1.28 & 78.60 ± 0.28 & 126.78 ± 2.87 & 3.78 ± 0.04 & 12.28 ± 0.25 & 14.16 ± 0.72 & 17.61 ± 0.45 \\
 &  & 31 & 9.02 ± 2.34 & 45.37 ± 1.73 & 54.50 ± 3.46 & 82.99 ± 8.53 & 3.05 ± 1.01 & 11.56 ± 0.77 & 12.74 ± 0.04 & 12.73 ± 0.44 \\
\cline{2-11}
 & \multirow[t]{3}{*}{2} & 3 & 14.13 ± 0.92 & 40.97 ± 1.63 & 72.47 ± 2.75 & 114.47 ± 3.38 & 7.47 ± 0.12 & 15.29 ± 0.36 & 22.25 ± 0.75 & 21.55 ± 1.34 \\
 &  & 16 & 11.03 ± 0.28 & 40.43 ± 2.01 & 66.75 ± 2.53 & 92.14 ± 9.73 & 5.29 ± 0.85 & 13.21 ± 1.31 & 21.45 ± 1.54 & 23.07 ± 2.62 \\
 &  & 31 & 5.57 ± 3.00 & 44.13 ± 2.10 & 59.30 ± 3.93 & 90.77 ± 7.56 & 3.16 ± 1.10 & 10.98 ± 1.52 & 19.87 ± 1.66 & 19.34 ± 1.43 \\
\cline{2-11}
 & \multirow[t]{3}{*}{3} & 3 & 15.07 ± 0.69 & \textbf{46.51} ± \textbf{1.19} & 77.09 ± 1.45 & 119.18 ± 1.28 & 7.87 ± 0.50 & \textbf{16.08} ± \textbf{0.57} & \textbf{26.33} ± \textbf{1.91} & 29.39 ± 1.75 \\
 &  & 16 & 8.75 ± 3.12 & 37.62 ± 2.64 & 67.83 ± 5.33 & 101.75 ± 7.19 & 4.11 ± 1.63 & 12.85 ± 1.86 & 22.33 ± 2.62 & 26.23 ± 1.37 \\
 &  & 31 & 1.78 ± 0.48 & 41.87 ± 0.60 & 59.19 ± 0.93 & 90.17 ± 2.70 & 1.12 ± 0.06 & 8.57 ± 0.75 & 20.88 ± 0.82 & 21.57 ± 1.40 \\
\cline{1-11} \cline{2-11}
\multirow[t]{9}{*}{1.0} & \multirow[t]{3}{*}{1} & 3 & 13.67 ± 0.14 & 37.53 ± 1.42 & 64.03 ± 2.69 & 101.80 ± 5.91 & 5.95 ± 0.35 & 8.04 ± 0.41 & 16.77 ± 1.28 & 17.16 ± 0.37 \\
 &  & 16 & 11.01 ± 0.78 & 43.18 ± 0.50 & 76.72 ± 3.71 & 124.57 ± 4.44 & 5.08 ± 0.48 & 8.81 ± 0.06 & 18.56 ± 1.06 & 21.14 ± 0.24 \\
 &  & 31 & 8.52 ± 2.84 & 38.78 ± 2.82 & 50.57 ± 4.03 & 80.32 ± 7.34 & 3.89 ± 1.49 & 8.52 ± 1.53 & 14.13 ± 1.18 & 16.03 ± 1.75 \\
\cline{2-11}
 & \multirow[t]{3}{*}{2} & 3 & 13.60 ± 0.09 & 38.81 ± 0.38 & 68.43 ± 1.45 & 110.24 ± 3.24 & 7.28 ± 0.14 & 11.81 ± 0.38 & 21.56 ± 1.07 & 29.45 ± 1.17 \\
 &  & 16 & 11.19 ± 0.70 & 37.77 ± 1.82 & 67.14 ± 0.59 & 97.45 ± 5.44 & 5.36 ± 0.59 & 10.46 ± 0.89 & 21.65 ± 0.75 & 27.57 ± 2.64 \\
 &  & 31 & 5.01 ± 2.29 & 42.19 ± 1.69 & 57.23 ± 1.44 & 91.31 ± 3.27 & 2.44 ± 1.38 & 7.34 ± 0.70 & 18.23 ± 1.35 & 25.42 ± 0.94 \\
\cline{2-11}
 & \multirow[t]{3}{*}{3} & 3 & 15.56 ± 0.63 & 44.49 ± 1.99 & 83.94 ± 5.12 & 124.95 ± 10.47 & \textbf{8.66} ± \textbf{0.47} & 14.82 ± 0.41 & 25.55 ± 1.58 & \textbf{41.52} ± \textbf{2.81} \\
 &  & 16 & 10.19 ± 0.99 & 34.81 ± 0.62 & 61.69 ± 6.84 & 96.79 ± 11.23 & 5.77 ± 0.66 & 10.67 ± 0.22 & 21.01 ± 1.08 & 32.52 ± 1.89 \\
 &  & 31 & -0.43 ± 1.00 & 39.15 ± 1.16 & 55.37 ± 1.21 & 86.81 ± 2.27 & 0.73 ± 0.76 & 4.82 ± 0.84 & 16.08 ± 1.52 & 28.45 ± 0.53 \\
\cline{1-11} \cline{2-11}
\multirow[t]{9}{*}{1.5} & \multirow[t]{3}{*}{1} & 3 & 14.60 ± 0.63 & 42.49 ± 1.20 & 78.85 ± 1.83 & 120.65 ± 1.90 & 5.81 ± 0.44 & 11.77 ± 0.63 & 10.25 ± 0.53 & 16.53 ± 0.39 \\
 &  & 16 & 12.31 ± 0.59 & 42.39 ± 2.04 & 80.93 ± 1.06 & 119.73 ± 4.15 & 4.23 ± 0.11 & 10.71 ± 0.25 & 9.13 ± 0.42 & 14.97 ± 0.77 \\
 &  & 31 & 9.91 ± 1.60 & 39.66 ± 1.57 & 50.67 ± 3.78 & 74.19 ± 5.61 & 3.39 ± 0.95 & 10.63 ± 0.28 & 8.37 ± 0.28 & 12.15 ± 0.47 \\
\cline{2-11}
 & \multirow[t]{3}{*}{2} & 3 & 14.71 ± 0.07 & 46.23 ± 1.40 & \textbf{87.14} ± \textbf{2.18} & \textbf{130.92} ± \textbf{3.49} & 7.32 ± 0.33 & 14.17 ± 0.22 & 18.81 ± 0.78 & 30.29 ± 1.30 \\
 &  & 16 & 11.90 ± 0.31 & 43.29 ± 0.81 & 80.72 ± 1.51 & 113.67 ± 2.34 & 5.76 ± 0.40 & 12.93 ± 0.81 & 17.91 ± 0.34 & 28.65 ± 1.18 \\
 &  & 31 & 7.98 ± 4.04 & 42.45 ± 1.96 & 65.91 ± 1.29 & 93.65 ± 2.56 & 3.43 ± 2.33 & 9.55 ± 2.46 & 14.42 ± 1.42 & 23.75 ± 0.43 \\
\cline{2-11}
 & \multirow[t]{3}{*}{3} & 3 & \textbf{16.00} ± \textbf{0.60} & 45.13 ± 0.39 & 81.44 ± 5.60 & 119.15 ± 12.59 & 7.38 ± 0.19 & 16.08 ± 1.50 & 23.91 ± 0.74 & 36.15 ± 3.95 \\
 &  & 16 & 9.92 ± 2.05 & 36.69 ± 2.42 & 60.16 ± 5.45 & 73.66 ± 10.08 & 4.92 ± 0.67 & 12.69 ± 1.18 & 18.31 ± 1.35 & 27.69 ± 3.21 \\
 &  & 31 & 5.58 ± 4.27 & 39.39 ± 5.56 & 60.33 ± 6.62 & 86.89 ± 7.94 & 2.69 ± 2.18 & 7.43 ± 3.84 & 14.77 ± 3.07 & 25.43 ± 3.68 \\
\cline{1-11} \cline{2-11}
\bottomrule
\end{tabular}
} 
\caption{Extensive sweep over number of agents, number of episode steps (relative to number of steps that would be required by a single agent to perform a scan of the entire board) and observation size. Average of 3 training seeds.}
\label{tab:ma_board_scans_eval_randomize_map_shap_map_width}
\end{table*}

% \subsection{Impact Statement}

% This paper presents work whose goal is to advance the field of Machine Learning. There are many potential societal consequences of the work, none which we feel must be specifically highlighted here.

% \section{Acknowledgments}
% Fuck the haters and the opps.

\bigskip

\bibliography{ref}

\end{document}